# Traffic Signs Detection and Recognition System using Deep Learning

Pavly Salah Zaki
Computer and Communication Department
Faculty of Engineering, Helwan University
Cairo, Egypt
pavly.salah.zaki@gmail.com

Marco Magdy William
Computer and Communication Department
Faculty of Engineering, Helwan University
Cairo, Egypt
marcomagdy241@gmail.com

Bolis Karam Soliman
Computer and Communication Department
Faculty of Engineering, Helwan University
Cairo, Egypt
boliskaram8@gmail.com

Kerolos Gamal Alexsan
Computer and Communication Department
Faculty of Engineering, Helwan University
Cairo, Egypt
kerolsgamal113@gmail.com

Maher Mansour
Faculty of Engineering, Helwan University
Cairo, Egypt
manmaher24@yahoo.com

Magdy El-Moursy
Mentor, A Siemens Business
Cairo, Egypt
magdy_el-moursy@mentor.com

Kerolos Khalil
Mentor, A Siemens Business
Cairo, Egypt
keroles_khalil@mentor.com

*Abstract*—With the rapid development of technology, automobiles have become an essential asset in our day-to-day lives. One of the more important researches is Traffic Signs Recognition (TSR) systems. This paper describes an approach for efficiently detecting and recognizing traffic signs in real-time, taking into account the various weather, illumination and visibility challenges through the means of transfer learning. We tackle the traffic sign detection problem using the state-of-the-art of multi-object detection systems such as *Faster Recurrent Convolutional Neural Networks* (F-RCNN) and Single Shot Multi-Box Detector (SSD) combined with various feature extractors such as *MobileNet v1* and *Inception v2*, and also *Tiny-YOLOv2*. However, the focus of this paper is going to be *F-RCNN Inception v2* and *Tiny YOLO v2* as they achieved the best results. The aforementioned models were fine-tuned on the German Traffic Signs Detection Benchmark (GTSDB) dataset. These models were tested on the host PC as well as Raspberry Pi 3 Model B+ and the TASS PreScan simulation. We will discuss the results of all the models in the conclusion section.

## I. Introduction

With the rapid technological advancement, automobiles have become a crucial part of our day-to-day lives. This makes the road traffic more and more complicated, which leads to more traffic accidents every year. According to the Association for Safe International Road Travel (ASIRT) organization, about 1.3 million people die (including 1,600 children under 15 years of age!), and about 20-50 million are injured or disabled annually due to traffic accidents [1].

There are numerous reasons that lead to those horrifying numbers of road accidents: according to San Diego Personal Injury Law Offices, the leading causes for such traumatic accidents are distracted driving and speeding [2]. Hence, a serious and immediate action needed to be taken. Advanced Driver Assistant System (ADAS) aims to help in that matter. ADAS refer to high-tech in-vehicle systems that are designed to increase road safety by alerting the driver of hazardous road conditions. Examples of the crucial ADAS sub-systems are Lane Departure, Collision Avoidance, and Traffic Signs Recognition (TSR). Recently, Traffic Signs Recognition has become a hot and active research topic due to its importance; there are various difficulties presented to the drivers that hinder their ability to properly see the traffic signs. Some of those difficulties are: *lighting conditions*, *weathering conditions*, *presence of other objects*, and more as shown in fig. 1. Hence it was necessary to automate the traffic signs detection and recognition process efficiently.

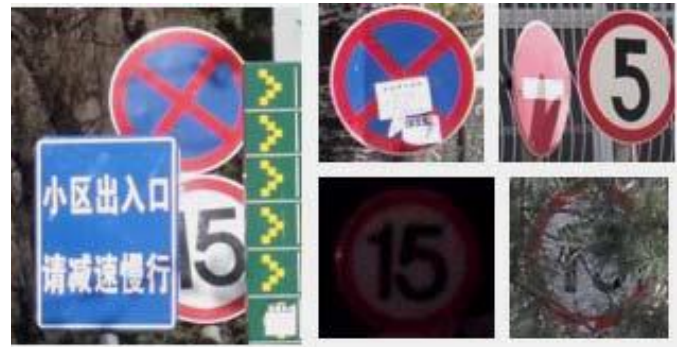

Fig. 1. Difficulties that may face TSR systems in real-life

According the German Traffic Signs Detection Benchmark (GTSDB) [3], Road traffic signs are divided into three main categories: **Prohibitory**, **Mandatory** and **Danger**. **Prohibitory** signs are used to ban certain behaviors, **Mandatory** signs indicate pedestrians, vehicles and intersections, and finally **Danger** signs alert drivers to be aware of dangerous targets on the road. These categories and their main defining features are shown in TABLE 1.

TABLE 1. Traffic signs categories according to GTSDB

| Category | Shape | Color | Example |
| --- | --- | --- | --- |
| **Prohibitory** | Circular | Red, Blue, White & Black | 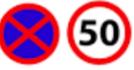 |
| **Mandatory** | Rectangular & Circular | Blue, White & Black | 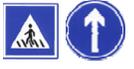 |
| **Danger** | Triangular | Red, White & Black | 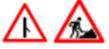 |

In this paper, a deep learning approach was taken because a model can learn the features from images autonomously from the training samples.

This paper is organized as follows: In *section II*, some previous related work is presented. The network structure and its implementation are described in details in *section III*. In *section IV*, experimental results for the proposed algorithm are provided. Finally, the paper is concluded with our personal notes regarding the system in *section V*.

## II. RELATED WORK

The study of traffic signs recognition started as the Program for European Traffic with High Efficiency and Unreasonable Safety (PROMETHEUS) funded by automobile companies such as Mercedes Benz in order to study traffic sign recognition system [4]. The traffic sign recognition consists of three stages—***detection***, ***tracking*** and ***recognition*** [5] as shown in fig. 2.

### A. Detection
The goal of the detection phase is to locate the regions of interest (RoI) in which the object is most likely to be found and indicate the object's presence. During this phase the image is segmented [6], a potential object is then proposed according to previously provided attributes such as color and shape.

### B. Tracking
In order to assure the correctness of the proposed region, a tracking phase is needed. Instead of detecting the image using only one frame, the algorithm would track the proposed object for a certain number of frames (usually four). This has proven to increase the accuracy significantly. The most common object tracker is the Kalman Filter [7].

### C. Recognition
The recognition phase is the main phase in which the sign is classified to its respective class. Older object recognition techniques may include statistical-based methods, Support Vector Machine, Adaboost and Principal Component Analysis [8].

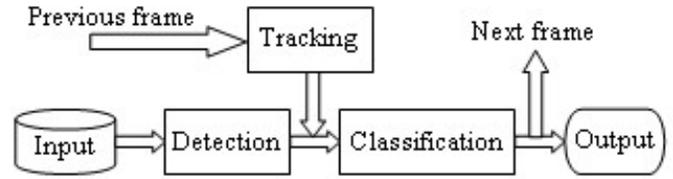

Fig. 2. Procedure of object detection

In the more recent years, deep learning approaches have become more and more popular and efficient. Convolutional Neural Networks (CNNs) [9] have achieved great success in the field of image classification and object recognition. Unlike the traditional methods, CNNs can be trained to automatically extract features and detect the desired objects significantly faster and more reliable [10].

## III. TRAFFIC SIGNS DETECTION AND RECOGNITION

### A. Dataset
Training and testing a Deep Convolutional Neural Network requires a large amount of data as a base. The German Traffic Sign Detection Benchmark (GTSDB) has become the de facto of training Deep CNNs when it comes to traffic sign detection. It includes many types of traffic signs in extreme conditions—weathering, lightening, angles, etc… which help the model train to recognize the signs found in those conditions. The GTSDB contains a total of 900 images (800 for training and 100 for testing). However, this number is clearly not enough for large-scale DCNN models such as F-RCNN Inception.

### B. System Design
- *Training phase*

First, the training images are loaded in RGB mode, they are then converted to HSV color space. Each image is then passed to the neural network for training. Finally, the network predicts where the traffic sign is (RoI extraction) followed by non-maximum suppression to choose only the RoIs with the highest confidence, then the model predicts to which class the signs belong. These predictions are then compared to the ground-truth (actual) regions of interest and class labels. The loss function is computed i.e. how far the model was from the correct prediction and back-propagation is then applied to decrease the loss value.

This process is repeated for a selected number of epochs, after that the training phase is said to be finished.

- *Testing phase*

In the testing phase, the images (or video frames) are loaded in RGB mode and then converted to HSV as well, but there is no training, the model just predicts the location and class of the sign as shown in fig. 3.

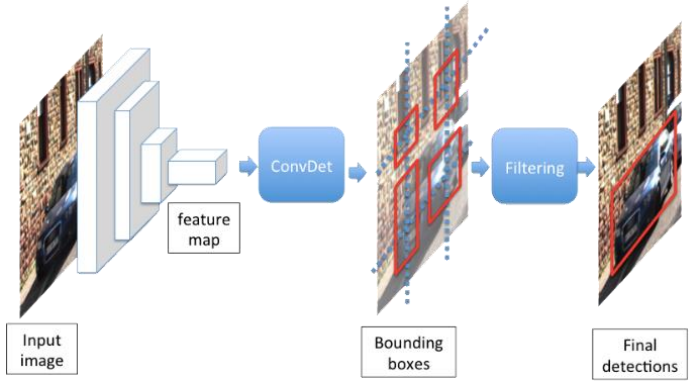

Fig. 3. Testing the model

## C. Network Structure

Various models have been trained and tested, but in this section, the F-RCNN Inception v2 and YOLO v2 models are presented since they produced the best overall results.

### I. Faster RCNN Inception v2

The first model is the Inception v2 [11] model as a front-end network structure of the Faster Recurrent Convolutional Neural Network (F-RCNN) [12] algorithm to detect and classify traffic signs. F-RCNN consists of Fast R-CNN detector and a Region Proposal Network (RPN) and then Non-Maximum Suppression is applied to choose the best region.

Equation (1) shows the method of calculating the Intersection over Union (IoU) in RPN to determine whether the proposed region contains an object or not. The idea is that we want to compare the ratio of the area where the two boxes *overlap* to the total *combined* area of the predicted and ground-truth boxes. If the value of the IoU is over the threshold of 0.7, that area is considered to be an object.

$$IoU = \frac{A \cap Gt}{A \cup Gt} \begin{cases} > 0.7 = \text{object} \\ < 0.3 = \text{not object} \end{cases} \quad (1)$$

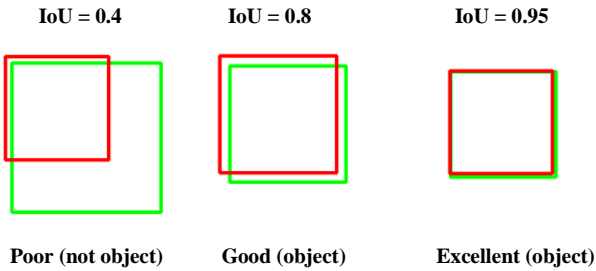

| IoU = 0.4 | IoU = 0.8 | IoU = 0.95 |
|---|---|---|
| Poor (not object) | Good (object) | Excellent (object) |

Fig. 4. Comparison between IoUs

TABLE 2 shows the network structure for the Inception v2 model. It consists mainly of 3x3 convolution (conv.) layers alongside 1x1 convolutions as they were proven to be effective in dimensionality reduction and thus faster performance. The base network consists of six conv. layers and a pooling layer. It is then followed by three times the network shown in fig. 5, five times the network shown in fig. 6 and two times the network shown in fig. 7. For classification a Softmax classifier is used.

TABLE 2. Inception v2 structure

| type | patch size/stride or remarks | input size |
|---|---|---|
| conv | 3×3/2 | 299×299×3 |
| conv | 3×3/1 | 149×149×32 |
| conv padded | 3×3/1 | 147×147×32 |
| pool | 3×3/2 | 147×147×64 |
| conv | 3×3/1 | 73×73×64 |
| conv | 3×3/2 | 71×71×80 |
| conv | 3×3/1 | 35×35×192 |
| 3×Inception | As in figure 5 | 35×35×288 |
| 5×Inception | As in figure 6 | 17×17×768 |
| 2×Inception | As in figure 7 | 8×8×1280 |
| pool | 8 × 8 | 8 × 8 × 2048 |
| linear | logits | 1 × 1 × 2048 |
| softmax | classifier | 1 × 1 × 1000 |

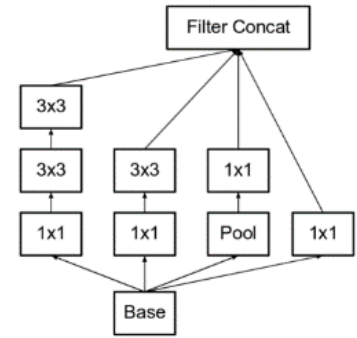

Fig. 5. Block 1 (used 3x in the Inceptionv2 architecture)

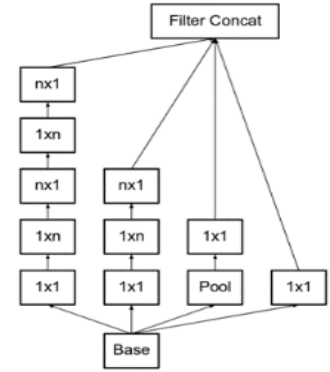

Fig. 6. Block 2 (used 5x in the Inceptionv2 architecture)

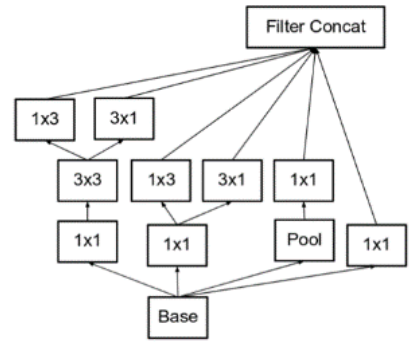

Fig. 7. Block 3 (used 2x in the Inceptionv2 architecture)

## II. Tiny-YOLO v2

The second used model is You Only Look Once (YOLO). YOLOv1 [13] is a state-of-the-art, real-time object detection system. On a Titan X it processes images at 40-90 FPS and has a mAP on VOC 2007 of 78.6% and a mAP of 48.1% on COCO test-dev. YOLO v2 is Better, Faster and Stronger than YOLO v1 [14].

It looks at the whole image at test time so its predictions are informed by global context in the image. It also makes predictions with a single network evaluation unlike systems like R-CNN which require thousands for a single image. This makes it extremely fast, more than 1000x faster than R-CNN and 100x faster than Fast R-CNN [14]. Fig. 8 shows some improvements of YOLOv2 over YOLOv1

| | YOLO | | | | | | | | YOLOv2 |
|---|---|---|---|---|---|---|---|---|---|
| batch norm? | | ✓ | ✓ | ✓ | ✓ | ✓ | ✓ | ✓ | ✓ |
| hi-res classifier? | | | ✓ | ✓ | ✓ | ✓ | ✓ | ✓ | ✓ |
| convolutional? | | | | ✓ | ✓ | ✓ | ✓ | ✓ | ✓ |
| anchor boxes? | | | | ✓ | ✓ | | | | |
| new network? | | | | | ✓ | ✓ | ✓ | ✓ | ✓ |
| dimension priors? | | | | | | ✓ | ✓ | ✓ | ✓ |
| location prediction? | | | | | | ✓ | ✓ | ✓ | ✓ |
| passthrough? | | | | | | | ✓ | ✓ | ✓ |
| multi-scale? | | | | | | | | ✓ | ✓ |
| hi-res detector? | | | | | | | | | ✓ |
| VOC2007 mAP | 63.4 | 65.8 | 69.5 | 69.2 | 69.6 | 74.4 | 75.4 | 76.8 | **78.6** |

Fig. 8. Incremental improvements of YOLO v2

The new model structure shown in TABLE 3, shows the usage of 1x1 convolution layers which reduces the number of parameters significantly, which in turn makes the model much faster.

The YOLO v2 architecture can be visualized in reference [15], and the full details about each block can be viewed by hovering over that block.

TABLE 3. YOLOv2 structure

| Type | Filters | Size/Stride | Output |
|---|---|---|---|
| Convolutional | 32 | 3 × 3 | 224 × 224 |
| Maxpool | | 2 × 2/2 | 112 × 112 |
| Convolutional | 64 | 3 × 3 | 112 × 112 |
| Maxpool | | 2 × 2/2 | 56 × 56 |
| Convolutional | 128 | 3 × 3 | 56 × 56 |
| Convolutional | 64 | 1 × 1 | 56 × 56 |
| Convolutional | 128 | 3 × 3 | 56 × 56 |
| Maxpool | | 2 × 2/2 | 28 × 28 |
| Convolutional | 256 | 3 × 3 | 28 × 28 |
| Convolutional | 128 | 1 × 1 | 28 × 28 |
| Convolutional | 256 | 3 × 3 | 28 × 28 |
| Maxpool | | 2 × 2/2 | 14 × 14 |
| Convolutional | 512 | 3 × 3 | 14 × 14 |
| Convolutional | 256 | 1 × 1 | 14 × 14 |
| Convolutional | 512 | 3 × 3 | 14 × 14 |
| Convolutional | 256 | 1 × 1 | 14 × 14 |
| Convolutional | 512 | 3 × 3 | 14 × 14 |
| Maxpool | | 2 × 2/2 | 7 × 7 |
| Convolutional | 1024 | 3 × 3 | 7 × 7 |
| Convolutional | 512 | 1 × 1 | 7 × 7 |
| Convolutional | 1024 | 3 × 3 | 7 × 7 |
| Convolutional | 512 | 1 × 1 | 7 × 7 |
| Convolutional | 1024 | 3 × 3 | 7 × 7 |
| Convolutional | 1000 | 1 × 1 | 7 × 7 |
| Avgpool | | Global | 1000 |
| Softmax | | | |

However, according to the Darkflow official GitHub repository, it is recommended to train YOLOv2 (or YOLOv3) on a high-end GPU. For that reason, alongside the embedded system implementation, YOLO-Lite (or Tiny-YOLOv2) model was used instead.

### TINY-Yolov2

YOLO-LITE [16], a real-time object detection model developed to run on portable devices such as a laptop or cellphone lacking a Graphics Processing Unit (GPU). The model was first trained on the PASCAL VOC dataset then on the COCO dataset, achieving a mAP of 33.81% and 12.26% respectively. YOLO-LITE runs at about 21 FPS on a non-GPU computer. This speed is 3.8× faster than the fastest state of art model, SSD Mobilenetv1.

TABLE 4. shows a clear speed advantage for Tiny YOLOv2 over the rest, which is needed for the implementation on embedded systems (Raspberry Pi 3 Model B+) which will be discussed in the next section.

TABLE 4. Comparison between various YOLO model variations

| Model | Layers | FLOPS (B) | FPS | mAP | Dataset |
|---|---|---|---|---|---|
| YOLOv1 | 26 | not reported | 45 | 63.4 | VOC |
| YOLOv1-Tiny | 9 | not reported | 155 | 52.7 | VOC |
| YOLOv2 | 32 | 62.94 | 40 | 48.1 | COCO |
| YOLOv2-Tiny | 16 | 5.41 | 244 | 23.7 | COCO |
| YOLOv3 | 106 | 140.69 | 20 | 57.9 | COCO |
| YOLOv3-Tiny | 24 | 5.56 | 220 | 33.1 | COCO |

### D. Training the Models

The models are trained on 900 images from the GTSDB dataset and *four classes* — *Prohibitory*, *Mandatory*, *Danger* and *Stop*.

Figs. 9 and 10 show the exploratory data analysis done on the 43 classes present in the GTSDB, it is clear that there are many classes *(e.g. Speed Limit 20, Restriction Ends, Bend, School Crossing, Go Right* and *Go Left)* that have *less than 20 instances* in the dataset (highlighted in *red*) – which is not an adequate amount to train a deep CNN model at all. Other *classes (e.g. Speed Limit 60, Speed Limit 80, No Overtaking* and *Stop*) have **20 to 60 instances** (represented by the **orange** bars) which is still not enough. Finally, there are classes (e.g. *Speed Limit 30, Speed Limit 50* and *Giveway*) which are represented by the **green** bars have more than **60 instances**. For this reason – lack of sufficient training data in the GTSDB dataset – the team decided to use the four super-classes.
Full data analysis can be viewed in reference [17].

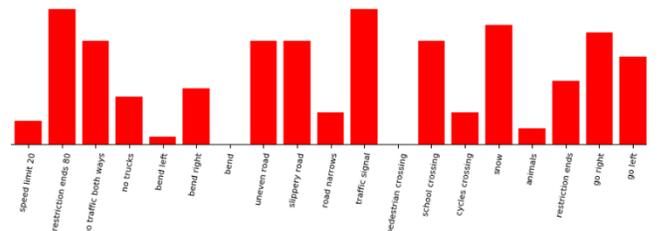

Fig. 9. Sample data visualization of traffic signs classes with low presence in the GTSDB (fewer than 20 instances)

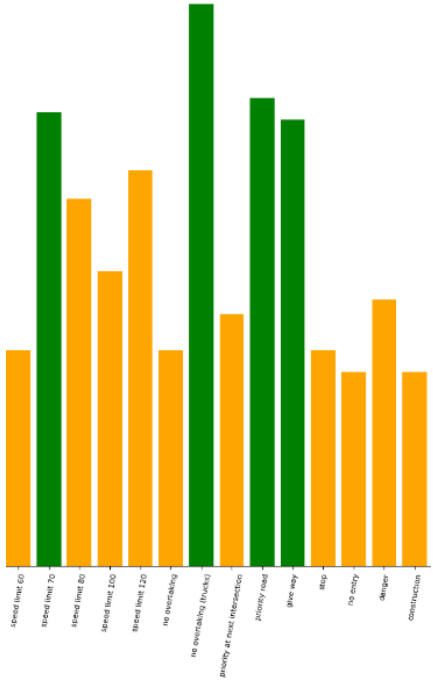

Fig. 10. Sample data visualization of classes with moderate (20 to 60 instances) and high (more than 60 instances) presence in the GTSDB

On the other hand, fig. 11 shows the data analysis on the four main classes used. There are no longer classes with less than 20 training samples, and most classes have more than 60 training samples (except for Stop which has 32).

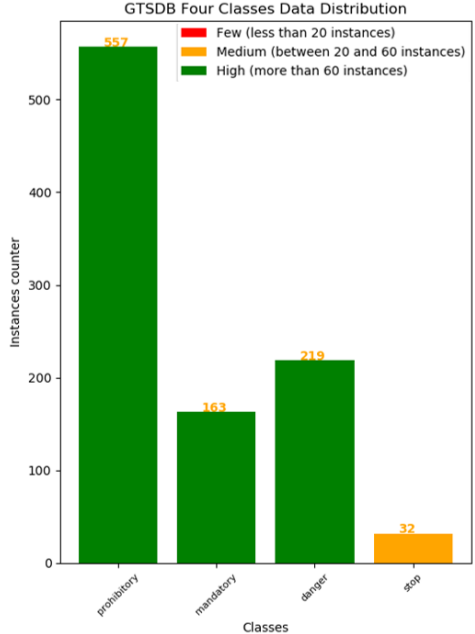

Fig. 11. Data visualization of the GTSDB on the 4 main classes

*Fine-tuning*

Fine-tuning is the process of re-training (i.e. doing back propagation) a model that was previously trained on a huge dataset like Microsoft COCO on a smaller dataset.

In order to get the best results, hyper-parameters need to be changed appropriately. For example, using data augmentation, normalization and regularization techniques, using drop-out, non-maximum suppression and learning rate decay and many more. Fine tuning is shown in TABLE 5.

Finally, the classifier needs to be changed according to the number of classes in the Dataset.

TABLE 5. Fine-tuning F-RCNN Inception v2 on the GTSDB

| Model | F-RCNN Inception v2 |
|---|---|
| Number of epochs | 8.5k |
| Data augmentation | Random grey scale<br>Random 25% crop<br>Random vertical flip<br>Random horizontal flip |
| Number of classes | 4 |
| Number of training samples | 900 |
| Fine-tuning | Resizer: fixed aspect ratio<br>Initial learning rate: 0.002<br>Learning rate decay: $1e^{-1}$ every 2,000 steps<br>IoU threshold: 0.7<br>Non-Max Suppression: 0.7<br>L2 Regularization<br>Batch Normalization<br>Drop out: 30%<br>Softmax classifier: 4 classes |

## IV. EXPERIMENTAL RESULTS

Fig. 12 shows some of the obtained results using our proposed the FRCNN Inception v2 model including false positives and false negatives.

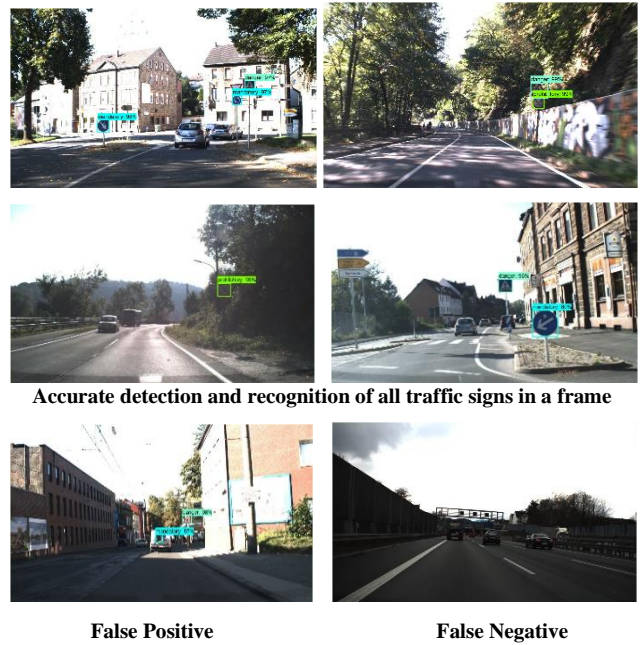

**Accurate detection and recognition of all traffic signs in a frame**

**False Positive**            **False Negative**

Fig. 12. Obtained results

TABLE 6 shows the performance (accuracy and speed) achieved by different models on the Host PC with a high-end GTX 1070 GPU on *720p* videos (and real-time video feed).

TABLE 6. Performance comparison on GTX 1070 GPU

| Model | mAP | Avg speed (FPS) |
|---|---|---|
| *SSD MobileNet v2* | 83% | 42 |
| *F-RCNN ResNet50* | 90% | ~20 |
| *F-RCNN Inception v2* | **96%** | ~25 |
| *Tiny-YOLO v2* | 73% | **~70** |

TABLE 7 shows the accuracies achieved by the F-RCNN Inception v2 and Tiny-YOLO v2 models on the four classes.

TABLE 7. F-RCNN Inception v2 and Tiny-YOLO v2 models achieved average accuracies on the four classes

| Model | Prohibitory | Mandatory | Danger | Stop |
|---|---|---|---|---|
| **F-RCNN Inception v2** | **98%** | **95%** | **96%** | **93%** |
| Tiny-YOLO v2 | 74% | 72% | 73% | 73% |

TABLE 8 shows the speeds (in FPS) achieved by the F-RCNN Inception v2 and Tiny-YOLO v2 models on various systems – GPUs and CPUs.

TABLE 8. Speed (FPS) comparison between different hosts operating the F-RCNN Inception v2 and Tiny-YOLOv2 models

| Host Specs | F-RCNN Inceptionv2 | Tiny-YOLOv2 |
|---|---|---|
| *CPU: I7 6500U @2.5GHz* | ~1 | **~18** |
| *GPU: GTX 1050 4GB* | ~13 | **~46** |
| *GPU: GTX 1070 6GB* | 25~30 | **65~70** |
| *GPU: Quadro P400 6GB* | ~20 | **45~55** |

To confirm that the system works on a low-end embedded system, the *SSD MobileNet v2 and Tiny-YOLOv2* (that were trained on four classes) were tested on the *Raspberry Pi 3 Model B+* produced an average speed of *2FPS* and *7FPS* respectively, which is enough for real-time applications. Result is shown in fig. 13.

Other models were tested as well. However, some models (e.g. FRCNN Inception v2) didn't even load properly and the process was 'killed' because the model was too heavy.

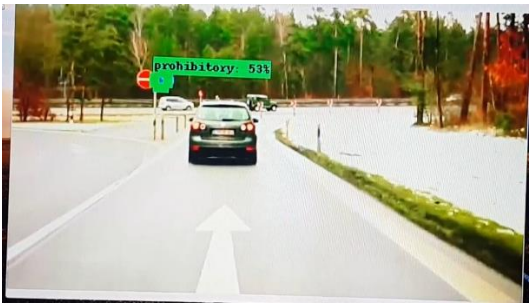

Fig. 13. Result on Raspberry Pi 3 Model B+ using the SSD MobileNetv2 model

Testing the FRCNN Inception v2 on the PreScan [18] simulation on a Quadro P4000 GPU achieved an average speed of 20 Frames Per Second.

TASS PreScan is a real-time self-driving car simulation on a life-like road containing pedestrians, traffic signs, traffic lights, various buildings, lighting conditions, weathering conditions etc… Results are shown in fig. 14.

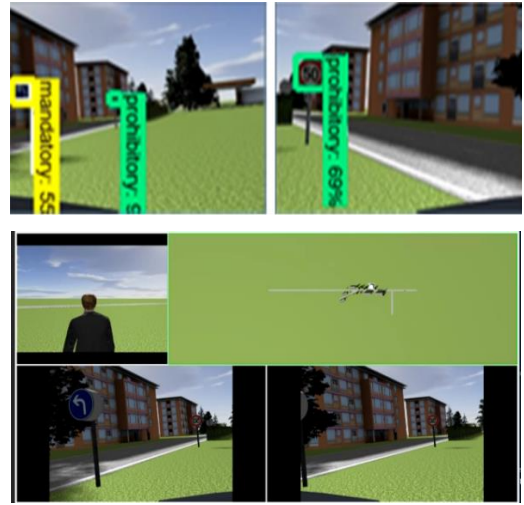

Fig. 14. Results on TASS PreScan simulation

TABLES 9 and 10 show the average accuracies and speeds achieved by the F-RCNN Inceptionv2 and Tiny-YOLOv2 models on the four classes vs Algorithm 1 which used Canny Edge Detector (for detection) and a CNN (for classification)* [19] and Algorithm 2 which used HCRE and SFC-tree method [20] respectively.

*This algorithm was tested on 5 traffic signs classes – *No Entry*, *Ahead Only*, *Turn Right*, *Turn Left* and *Ahead or Turn Tight Ahead*.

TABLE 9. Achieved average accuracies F-RCNN Inception v2 and Tiny-YOLO v2 models vs Algorithm 1 and Algorithm 2

| Model | F-RCNN Inceptionv2 | Tiny-YOLOv2 | Algorithm 1 | Algorithm 2 |
|---|---|---|---|---|
| *Accuracy* | **96%** | 73% | 87% | 95.76% |

TABLE 10. F-RCNN Inception v2 and Tiny-YOLO v2 models achieved average speeds vs Algorithm 1 and Algorithm 2

| Model | F-RCNN Inceptionv2 | Tiny-YOLO v2 | Algorithm 1 | Algorithm 2 |
|---|---|---|---|---|
| *GTX 1070* | 25~30 | **65~70** | 50 | 20 |
| *TASS PreScan (on Quadro P4000 GPU)* | ~20 | **45~55** | Not tested | Not tested |
| *Raspberry Pi 3 Model B+* | ~2 | **~7** | 15* | Not tested |

*Tested on a 480p video, whereas the rest are tested on 720p videos as mentioned before. Also, the algorithm was implemented on the Raspberry Pi 2.

## V. CONCLUSIONS

In this paper, we proposed a fast and effective method to detect and classify traffic signs. The main contributions of this paper are as follows:

- Using a fully convolutional network and transfer learning, the F-RCNN Inception v2 model has managed to achieve accurate, reliable and fast results even in complex real-life road situations (average of 96% accuracy).

- Tiny-YOLOv2 is a super-fast model with a decent accuracy, but if higher accuracy is needed, YOLOv2 or YOLOv3 should be used instead.

- After training the Inception v2 model on the GTSRB [21], on 39,200 images, 43 classes and using similar configuration as shown in section III-D, an accuracy of 99.8% was achieved – which is a record according to GTSRB competition [22].

- Accuracy improvements can be achieved by adding significantly more training data (at least 40k images, for an average of 1,000 images for each class) and training the models for a longer time if a high-end GPU is available.